\documentclass[letterpaper, 10 pt, conference]{ieeeconf} 
\IEEEoverridecommandlockouts                             
\usepackage{multicol}
\usepackage[bookmarksnumbered=true]{hyperref} 

\hypersetup{
     colorlinks = true,
     linkcolor = blue,
     anchorcolor = blue,
     citecolor = blue,
     filecolor = blue,
     urlcolor = blue
     }
\usepackage{amsmath, amssymb}
\usepackage{bbm}
\usepackage{comment}
\usepackage{subfigure}
\usepackage{mathtools}
\usepackage[ruled,linesnumbered,vlined]{algorithm2e}
\usepackage{algpseudocode}
\usepackage{graphicx}
\usepackage{cite}
\usepackage{array}
\usepackage{makecell}

\title{\LARGE \bf Object Rearrangement with Nested Nonprehensile Manipulation Actions}
\author{
Changkyu Song and Abdeslam Boularias$^{1}$
\thanks{$^{1}$The authors are with the Department of Computer Science of Rutgers University, Piscataway, New Jersey 08854, USA.
        {\tt\footnotesize \{cs1080, ab1544\}@cs.rutgers.edu}}%
}

\begin{document}
\maketitle
\thispagestyle{empty}
\pagestyle{empty}

\begin{abstract}
This paper considers the problem of rearrangement planning, i.e finding a sequence of manipulation actions that displace multiple objects from an initial configuration to a given goal configuration. Rearrangement is a critical skill for robots so that they can effectively operate in confined spaces that contain clutter. Examples of tasks that require rearrangement include packing objects inside a bin, wherein objects need to lay according to a predefined pattern. In tight bins, collision-free grasps are often unavailable. Nonprehensile actions, such as pushing and sliding, are preferred because they can be performed using minimalistic end-effectors that can easily be inserted in the bin. Rearrangement with nonprehensile actions is a challenging problem as it requires reasoning about object interactions in a combinatorially large configuration space of multiple objects.
This work revisits several existing rearrangement planning techniques and introduces a new one that exploits nested nonprehensile actions by pushing several similar objects simultaneously along the same path, which removes the need to rearrange each object individually.
Experiments in simulation and using a real Kuka robotic arm show the ability of the proposed approach to solve difficult rearrangement tasks while reducing the length of the end-effector's trajectories.

\end{abstract}

\IEEEpeerreviewmaketitle

\section{Introduction}
\begin{figure}[t]
\centering
\includegraphics[width=0.239\textwidth]{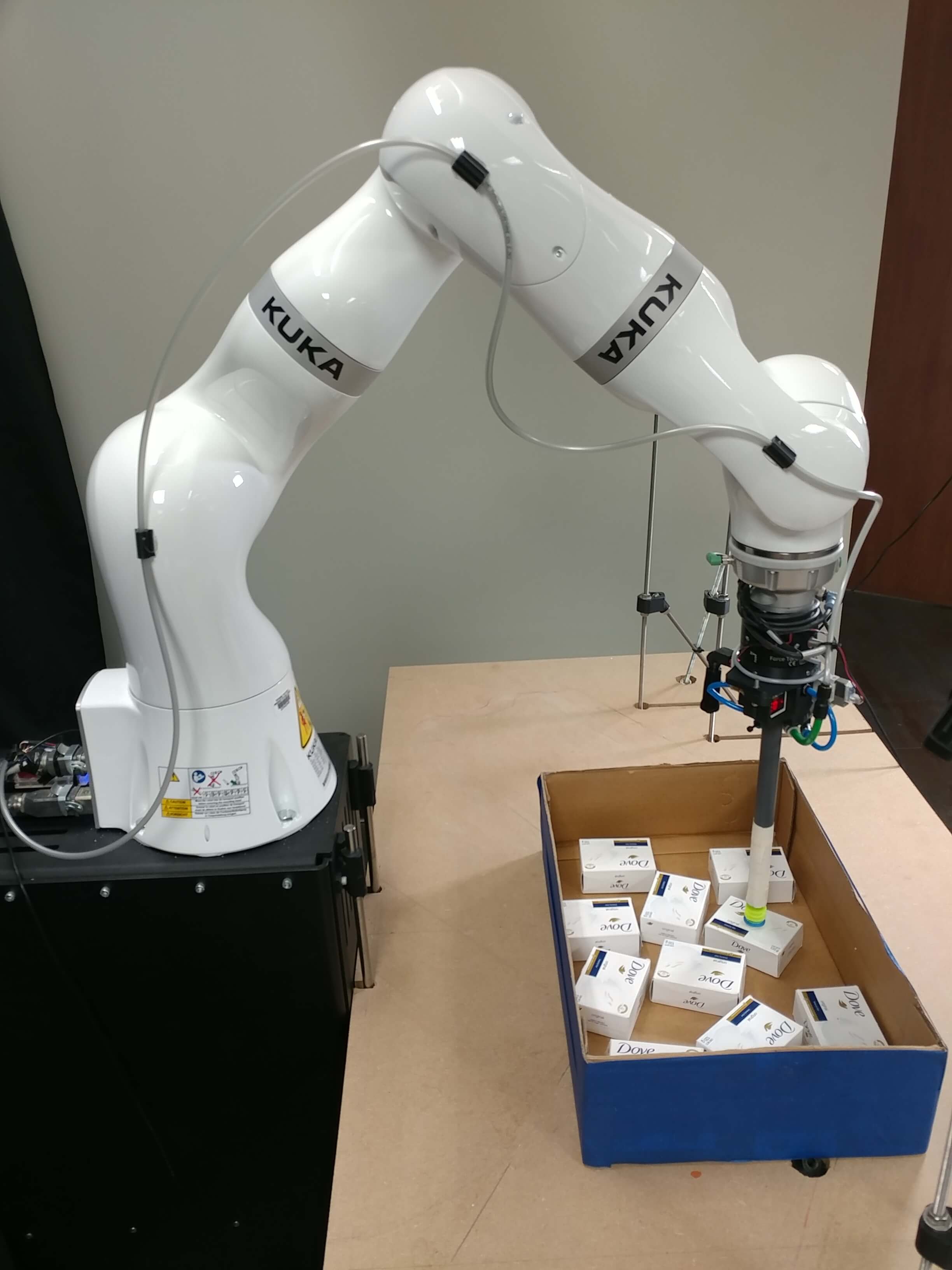}
\includegraphics[width=0.239\textwidth]{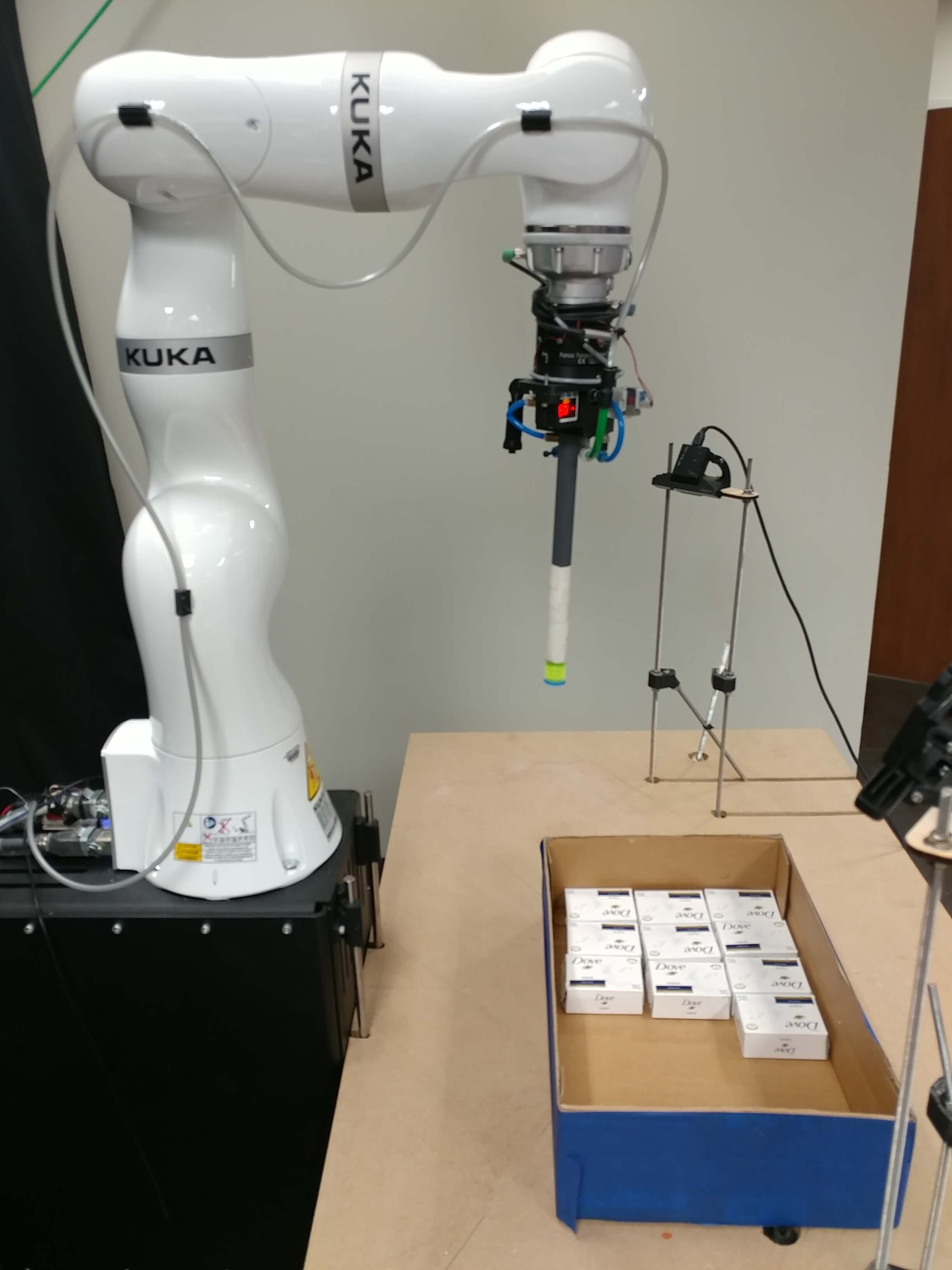}
\caption{The product packing problem for cuboid products: initial configuration (left), and achieved goal configuration (right).
Experiments are performed using a {\it Kuka} robotic arm, equipped with a minimalistic end-effector and a depth-sensing camera {\it SR300}.}
\label{fig:problem_bins}
\end{figure}

Warehouse automation has witnessed a vast growth in the past few years thanks to the deployment of intelligent robotic solutions. The {\it Amazon Kiva} mobile robotic fulfillment system is an example of the impressive progress that has been recently achieved in this area. Nevertheless, a large portion of tedious labor performed in warehouse operations is still manual to a large extent. For example, picking and packing products inside bins, such as the small shipping box in Figure~\ref{fig:problem_bins}, is a task that remains highly challenging for robotic systems. In addition to dealing with vision-related challenges, a robotic system needs to plan the trajectories of its arm and the actions to perform by reasoning about the effects of the collisions between the objects in the scene. Moreover, bin packing requires reasoning about the different orders in which the objects should be moved to their marked final poses, which is a combinatorial optimization problem known as object rearrangement planning.

Solving rearrangement problems optimally is computationally prohibitive. Therefore, most proposed algorithms content with finding feasible plans~\cite{Krontiris-RAS-2014,Krontiris-RSS-15,7487581}. The few efficient and optimal algorithms that have been proposed require additional constraints, such as having a clear buffer zone for intermediate placement of objects~\cite{RSS172,HanStiBekYu18RAL}. Moreover, most of these works use only {\it pick-and-place} actions and do not take advantage of nonprehensile actions, such as pushing and pulling, that can solve these tasks more efficiently and using only minimalistic end-effectors.
While nonprehensile manipulation is a growing topic in robotics~\cite{Dogar2011AFF,GuanVR18,Barry2012,Jentzsch2015,Pinto-abs-1810-10654}, it has not yet been tested on real and dense rearrangement tasks such as the one in Figure~\ref{fig:problem_bins}. In this paper, we present and empirically evaluate an approach for solving this type of tasks. The presented approach utilizes nested pushing actions for moving several objects simultaneously to their targets whenever possible. While global optimality is not a criterion that is maintained or quested by the present work, the proposed approach is tailored for efficient bin packing applications. At a high level, a tree search with backtracking is used for finding subsets of objects to push together. At an intermediate level, trajectories of objects are merged at optimized rendezvous points to benefit from grouped pushing actions. At a low level, the problem of pushing objects to local target points is formulated as a Markov Decision Process and solved offline. The obtained policy is stored in a look-up table and utilized online by the high level planner. Extensive experiments in simulated environments as well as on the real rearrangement tasks shown in Figure~\ref{fig:problem_bins} clearly demonstrate the effectiveness of this approach.

\section{Related Work}
Object rearrangement using non-prehensile actions is related to several areas in robotic manipulation and planning.

{\it Manipulation planning:}
Object rearrangement problems can be solved with generic sampling-based motion planning algorithms, such as RRT$^*$~\cite{Karaman:2011}, Kinodynamic RRT~\cite{WebbB13,king2016}, and trajectory optimization algorithms such as CHOMP~\cite{Zucker2013CHOMPCH}. However, these algorithms do not scale up to spaces that result from combining the configuration spaces of a robot and several manipulated objects. Similar scalability issues are encountered in multi-robot motion planning due to the large number of degrees of freedom~\cite{Solovey2014FindingAN}. 
In the context of object grasping, the dimension of the configuration is typically
reduced by imposing various motion constraints relative to {\it transfer} and {\it transit} actions~\cite{Mirabel7989462}. It is also often common to explicitly introduce intermediate actions, such as {\it reconfiguration}, for cost and computation efficiency~\cite{King2013PregraspMA}. Our method is related to
the probabilistic path planning method for multiple robots with subdimensional expansion~\cite{Wagner2012ProbabilisticPP}, wherein plans in each individual robotâs configuration space are obtained separately, then those spaces are entangled when robots come into close proximity with one another. In our method, individual spaces of objects are coupled when the objects are physically close to each other.

{\it Object Rearrangement using Prehensile Actions:} Most prior works on objects rearrangement focused on {\it pick-and-place} actions for transporting objects on collision-free paths~\cite{Krontiris-RAS-2014}. One proposed algorithm adapts a manipulation approach for clearing a path to an unreachable object~\cite{Stilman2007ManipulationPA}, and uses a tree-search with backtracking for finding a feasible order in which objects can be picked and placed~\cite{Krontiris-RSS-15}. The same algorithm was adapted for solving non-monotone rearrangement tasks, by temporarily moving obstacles to buffer zones. Our approach borrows the general idea of tree-based search from this line of works.
A more efficient algorithm that avoids backtracking builds a directed acyclic dependency graph that indicates which object is on the way of which other object, and the objects are then manipulated in the topological order of the dependency graph~\cite{7487581}. Tabletop rearrangement with overhand grasps was also formulated as a TSP problem and an algorithm that minimizes the total distance traveled by the end-effector was proposed~\cite{RSS172,HanStiBekYu18RAL}. 
Rearrangement planning has also been used in robotic assembly of discrete architectural structures~\cite{DBLP:journals/corr/abs-1810-00998}.

{\it Nonprehensile Manipulation:}
Treating objects always as obstacles that must be avoided at all costs often results in inefficient plans. In practice, collisions can be safe and even helpful in many cases. For example, combined pushing and grasping actions have been shown to succeed where traditional grasp planners fail, and they work well under uncertainty by using the funneling effect of pushing~\cite{Dogar2011AFF}. 
Environmental contact and compliant manipluation were also leveraged in peg-in-hole planning tasks~\cite{GuanVR18}. 
As in our rearrangement approach, interactions between objects and end-effectors in~\cite{GuanVR18} are modeled using an MDP near contact manifolds only, while sample-based planners are used in the other regions. DARRT is a sampling-based algorithm for general-purpose motion planning problems with diverse, non-prehensile manipulation actions~\cite{Barry2012}, based on the RRT structure~\cite{Jentzsch2015}. Our approach also uses RRT and non-prehensile manipulation actions, but it is tailored for efficient rearrangement. The problem of pushing a single target object to a desired goal region was also recently solved through reinforcement learning~\cite{Pinto-abs-1810-10654}. In the present work, we are concerned with nested pushing actions, where one or multiple objects are pushed simultaneously.

{\it Object Rearrangement with Nonprehensile Actions:}
A framework that plans rearrangement of clutter using non-prehensile actions, such as pushing, was introduced by
Dogar et al.~\cite{Dogar2012}, but it did not consider nested actions. A version of RRT with Kinodynamics also used nonprehensile whole arm rearrangement planning~\cite{king2015,king2016}. This approach is related to ours in the sense that whole arm manipulation can be seen as a special type of nested manipulation actions. Monte Carlo tree-search was used to solve rearrangement problems in clutter, with simple non-prehensile actions, the search was guided by policies learned from demonstrations~\cite{KingICRA2017}. A similar problem was solved with deep Q-learning~\cite{DBLP:conf/icra/YuanSKWH18}.

{\it Task and Motion Planning:} Rearrangement is a special instance of the general {\it Task and Motion Planning} (TMP) problem~\cite{Srivastava2014CombinedTA}. 
For example, an iterative TMP algorithm was demonstrated on rearrangement tasks, with the constraints on motion feasibility removed at the task level~\cite{Dantam2016IncrementalTA}. Other TMP algorithms rely on factored transition functions~\cite{GarrettRSS17}, 
impulse exchange constraints at the path optimization
level~\cite{18-toussaint-RSS}, or symbolic representations of the preconditions and effects of the manipulation actions~\cite{Konidaris2014ConstructingSR}. Efficient heuristics for TMP have also been investigated~\cite{Garrett2014FFRobAE,doi:10.1177/0278364917739114}.

{\it Navigation and Manipulation among Movable Objects (NAMO):}
Rearrangement problems are related to NAMO~\cite{DBLP:journals/ar/StilmanNKK07}, which is known to be NP-hard even for simple instances~\cite{DEMAINE200321}. Thus, most efforts have focused on finding time-efficient algorithms that are near-optimal~\cite{Stilman2007ManipulationPA,doi:10.1177/0278364913507795,Levihn2012HierarchicalDT}. Minimum constraint removal is another problem related to NAMO, where the planner searches for a transfer path that collides with the minimum number of objects~\cite{Hauser2012TheMC,Hauser2013MinimumCD}. NAMO techniques were used to improve the computational efficiency of rearrangement planning~\cite{7487581}, and can also be used for the approach proposed in the current work.



\section{Problem Setup and Notation}
Consider a 3D workspace that contains:
\begin{itemize}
    \item a set of static obstacles $\mathcal{S}$,
    \item a set of $n$ movable rigid-body objects $\mathcal{O} = \{o_1, o_2, \dots, o_n\}$ on a tabletop, where each object $o_i\in\mathcal O$ is described by its $3D$ pose (2D position and rotation angle) $p[i]\in \mathcal{P}_i \subseteq SE(2)$,
    \item and a robotic arm able to rotate, move and push objects. The configuration of the arm is denoted by $q\in \mathcal Q$. 
\end{itemize}

Configuration space $\mathcal{C}$ is defined as the Cartesian product of the configuration spaces of the robot and objects: $\mathcal{C} = \mathcal{Q} \times \prod_{i=1}^{n} \mathcal{P}_i$.
The valid subset of $\mathcal{C}$ contains all configurations where no object is partly or entirely inside another object, including the robot's arm, and no object is colliding with a static obstacle. Rigid-body collisions between the movable objects, or between the arm and the movable objects, are permitted and used for rearrangement. 
We denote by $q(p)$ a collision-free configuration of the robotic arm when the end-effector is perpendicular to the tabletop and centered in the 2D position position given by pose $p$. 

There are three different types of valid configurations: {\it i) Stable}: all the objects rest on the surface and the robotic arm is idle. {\it ii) Transit}: the robotic arm moves from an initial configuration $q_I$ to a final one $q_F$ while avoiding any collision with the objects, which are described by their joint pose $p$, we use {\tt TRANSIT}$(q_I,q_F,p)$ to describe a function that returns a collision-free path to this end. Sets of static obstacles $\mathcal{S}$ and of objects $\mathcal{O}$ are constant during planning and do not need to be provided as arguments to the different functions that we define here. {\it ii) Push}: the end-effector simultaneously pushes a subset of objects $\mathcal{O'}\subseteq \mathcal O$ from their initial joint poses in $p_I$ to final poses in $p_F$. Collisions with other movable objects are allowed during the transfer, except with the objects that have been successfully moved to their final poses, denoted by set $\mathcal{O}_s$. We use {\tt NESTED-PUSH}$(q_I,p_I,p_F,\mathcal{O'},L)$ to denote the corresponding path of the end-effector. $L$ is a sorted list that indicates the order in which the objects of $\mathcal{O'}$ should be pushed, i.e the first object in the list pushes the second, while the second pushes the third, and so on.
Following the formulation introduced in~\cite{Krontiris-RSS-15}, a legal mode change between two modes is one where the configuration of the robot and objects at the end of the first mode is the same as at the start of the second mode. 
A rearrangement path $\pi: [0,1]:\rightarrow \mathcal{C}^H$ is a sequence of states in the joint configuration space $\mathcal{C}$ with legal mode changes. We use $\pi(t).q$ to denote the configuration of the arm at time-step $t$ in path $\pi$, and $\pi(t).p[i]$ to denote the pose of object $i$.
Finally, the nonprehensile rearrangement problem is defined as the problem of finding a path $\pi$ given $x_I = (q_I,p_I)$ and $x_F = (q_F,p_F)$ so that $\pi(0)=x_I$ and $\pi(1)=x_F$. If we define $cost(\pi)$ as the length of the trajectory traversed by the end-effector, then the optimal rearrangement problem is defined as finding the path $\pi$ with the minimum cost $cost(\pi)$ and that solves the rearrangement task. 


\section{Proposed Approach}
We first present the general search algorithm for finding a feasible order in which the objects should be pushed. Then, we present the nested pushing approach. Finally, we explain how we obtain the low-level pushing policy. 
\subsection{Nested Rearrangement Search (NRS)}
The proposed high-level NRS algorithm is given in Algorithm~\ref{alg:nrs} and is an adaptation of an existing rearrangement planning algorithm~\cite{Krontiris-RSS-15}, which itself is an adaptation of a NAMO approach for clearing obstructed paths~\cite{Stilman2007ManipulationPA}. The original algorithm of~\cite{Krontiris-RSS-15} is limited to grasping actions, while the current algorithm considers pushing actions. Moreover, the NRS algorithm also considers nested actions 
that are applied on sets of objects, as opposed to the mono-object manipulation actions used in~\cite{Krontiris-RSS-15}. On the other hand, the algorithm of~\cite{Krontiris-RSS-15} deals also with non-monotone instances, while we focus in this work on monotone instances for simplicity's sake. The subroutines related to pushing and nested operations are presented in the subsequent sections. 

Algorithm~\ref{alg:nrs} is a recursive function that takes as input the set $\mathcal{O}_r$ of objects that have not yet been placed in their final poses. In the first call of the function, $\mathcal{O} = \mathcal{O}_r$. Objects that have been successfully placed in their final poses are treated as obstacles during planning, to avoid infinite cycles. Many real-world rearrangement tasks can be solved accordingly. For example, bin packing can be achieved by first placing objects that are close to the edge of the bin. Once the first row is formed, it can be treated as a static obstacle. The bin can be packed by repeating this process, row by row. 

Additionally, the algorithm receives the current pose $q_I$ of the robotic arm and the current joint pose $p_I$ of all the objects, which is generally different from the starting pose provided in the first call, because the objects are generally displaced as a result of collisions and pushing actions.  The final poses $(q_F,p_F)$ are fixed. The algorithm stops when all objects are correctly placed in their final poses (line 1), in which case a path $\pi$ that moves the end-effector to its final configuration $q_F$ while avoiding any collision is returned (lines 2-3). Otherwise, the algorithm proceeds into partitioning the remaining objects set $\mathcal{O}_r$ into its powerset of subsets $P$, with each subset corresponding to a different group of objects that will be potentially pushed simultaneously. Partitions $P$ include sets containing single objects. Sets containing more than one object are manipulated using nested pushes. Given a set $P$, we create a list of all possible permutations of the objects in $P$ by using subroutine {\tt NLC} given in Algorithm~\ref{alg:nlc}. The set of permutations is denoted by $\mathcal L$ (line 5). The algorithm then proceeds into searching for a permutation that provides a feasible nested pushing action that moves all objects in $P$ into their final poses (lines 7-10). Nested pushing actions are performed using subroutine {\tt NESTED-PUSH}, explained in the next section, which takes as inputs the current state $(q_I,p_I)$ as well as the final joint configuration $p_F$. An ordered list $L$ indicates the subset of objects in $\mathcal{O}$ that will need to be pushed as well as the order of the nested pushes. {\tt NESTED-PUSH} returns a path $\pi$ of the arm-objects configuration (line 8). If the corresponding action is feasible, then $\pi\neq \emptyset$ and the search of permutations stops. The same algorithm is called recursively after removing from $\mathcal{O}_r$ the objects that were successfully placed in the current call (line 12). If it is possible to place the remaining objects, then the algorithm stops after concatenating paths $\pi$ and $\pi'$ (line 14), otherwise another subset of objects $P$ is tested, until a feasible order of nested pushing actions of subgroups of objects is found. 
An empty path $\emptyset$ is returned if no feasible order is found.
\begin{algorithm}[h]
\KwIn{$\mathcal{O}_r$: set of objects that have not yet been moved to their final poses,  $(q_I,p_I)$: initial state, $(q_F,p_F)$: final state\;}
\KwOut{Path $\pi$ of the joint arm-objects configuration\;}
\If{$\mathcal{O}_r = \emptyset$}
    {
        $\pi \leftarrow${\tt TRANSIT}$(q_I,q_F,p_I)$\;
        return $\pi$\;
    }
\ForEach{$P \in \mathcal{P}(\mathcal{O}_r) \setminus \{\emptyset\} $}
{
$\mathcal{L} = ${\tt NLC}$(P)$\;
$\pi\leftarrow\emptyset$\;
\ForEach{$L \in \mathcal{L}$}
    {
    $\pi\leftarrow${\tt NESTED-PUSH}$(q_I,p_I,p_F,\mathcal{O}\setminus \mathcal{O}_r,L)$\;
    \If{$\pi\neq \emptyset$}
            {
                break\;
            }
    }
    \If{$\pi\neq \emptyset$}
    {
        $\pi' \leftarrow${\tt NRS}$(\mathcal{O}_r \setminus P,\pi(1).q,\pi(1).p,q_F,p_F)$\;
        \If{$\pi'\neq \emptyset$}
        {
            $\pi \leftarrow \{ \pi | \pi' \}$\;
            return $\pi$\;
        }
    }
}
       return $\emptyset$ \;

\caption{Nested Rearrangement Search ({\tt NRS})}
\label{alg:nrs}
\end{algorithm}

\begin{algorithm}[h]
\KwIn{$\mathcal{O}_c$: set of objects to chain\;}
\KwOut{$\mathcal{L}$: set of permutations of objects in $\mathcal{O}_c$ \;}
$\mathcal{L} \leftarrow \emptyset$ \;
\ForEach{$o \in \mathcal{O}_c$}
{
\ForEach{$L \in${\tt NLC}$(\mathcal{O}_c \setminus \{o\})$}
{
$\mathcal{L} \leftarrow \mathcal{L} \cup \{ (o,L)\}$\;
}
}
return $\mathcal{L}$;
\caption{Nested List Construction ({\tt NLC})}
\label{alg:nlc}
\end{algorithm}

\subsection{Nested Pushing}

\begin{algorithm}[th]
\KwIn{$(q_I,p_I)$: initial state, $p_F$: final configuration of the objects, $\mathcal{O}_s$: set of objects that are successfully placed in their final configurations, $L$: sorted list of objects to push\;}
\KwOut{Path $\pi$ of the joint arm-objects configuration\;}
$N \leftarrow size(L)$\;
\If{N = 0}{return $\emptyset$ \;}
$\pi \leftarrow \emptyset$; $m\leftarrow 1$\;
$\pi_{main}\leftarrow${\tt PUSH}$(q_I,p_I,L[N], p_F[L[N]],m,\mathcal{O}_s)$\;
\If{$\pi_{main}=\emptyset$}{return $\emptyset$ \;}
\For{$i \leftarrow N-1$;\ $i > 0$;\ $i \leftarrow i - 1$}{
   $m\leftarrow N-i+1$\;
   $p_{\textrm{docking,1}}\leftarrow${\tt DOCKING-POSE}$(\pi_{main},p_I, L,m)$\;
   $p_{\textrm{docking,2}}\leftarrow${\tt NEAREST-POSE}$(\pi_{main},p_I,L,m,i)$\;
   $p_{\textrm{rdv}}\leftarrow${\tt RDV-POSE}$(\pi_{main}, L,m,p_{\textrm{docking,2}})$\;
   $\pi_1\leftarrow${\tt PUSH}$(q_I,p_I,L[i],p_{\textrm{docking,1}},1,\mathcal{O}_s)$\;
   $\pi_2\leftarrow${\tt PUSH}$(\pi_1(1).q,\pi_1(1).p,L[N],p_{\textrm{rdv}},m,\mathcal{O}_s)$\;
   $\pi_3\leftarrow${\tt PUSH}$(q_I,p_I,L[N],p_{\textrm{rdv}},m,\mathcal{O}_s)$\;
   $\pi_4\leftarrow${\tt PUSH}$(\pi_3(1).q,\pi_3(1).p,L[i],p_{\textrm{docking,2}},m,\mathcal{O}_s)$\;
       \If{$(\pi_1=\emptyset \vee \pi_2=\emptyset) \wedge (\pi_3=\emptyset \vee \pi_4=\emptyset) $}{return $\emptyset$ \;}
     \eIf{$cost(\{\pi_1 | \pi_2\}) \leq cost(\{\pi_3 | \pi_4\})$}
        {$\pi\leftarrow \{\pi | \pi_1 | \pi_2\}$;$(q_I,p_I) \leftarrow (\pi_2(1).q,\pi_2(1).q)$;}
        {$\pi\leftarrow \{\pi | \pi_3 | \pi_4\}$;$(q_I,p_I) \leftarrow (\pi_4(1).q,\pi_4(1).q)$;}
   }
   $\pi_{all}\leftarrow${\tt PUSH}$(\pi(1).q,\pi(1).p,L[N], p_F[L[N]],N,\mathcal{O}_s)$\;
   $\pi\leftarrow \{\pi | \pi_{all}\}$\;
   \For{$i \leftarrow N-1$;\ $i > 0$;\ $i \leftarrow i - 1$}
   {
      $\mathcal{O}_s\leftarrow \mathcal{O}_s \cup \{L[i+1]\}$\;
      $\pi_{i}\leftarrow${\tt PUSH}$(\pi(1).q,\pi(1).q,L[i], p_F[L[i]],1,\mathcal{O}_s)$\;
      \If{$\pi_{i}=\emptyset$}{return $\emptyset$ \;}
      $\pi\leftarrow \{\pi | \pi_{i}\}$\;
   }
   return $\pi$ \;
\caption{Nested Pushing ({\tt NESTED-PUSH})}
\label{alg:np}
\end{algorithm}

Algorithm~\ref{alg:np} returns a path $\pi$ that moves simultaneously a set of objects to a given goal configuration. The algorithm receives as inputs an initial state $(q_I,p_I)$, a goal configuration of the objects $p_F$, a list $L$ indicating the objects that will be pushed and the order in which the will be nested, in addition to the set of objects that are already successfully placed, to avoid colliding with them. If the list of objects is empty, then the algorithm returns an empty path (line 3). The returned path $\pi$ is initialized to an empty list. Counter $m$ counts the number of nested objects that are successfully lined up in front of the end-effector and are ready to be pushed to their final poses. While the algorithm described here is general, we found from our real bin packing experiments that it is practical only for $m\leq 2$.
First, the algorithm finds a path $\pi_{main}$ for pushing the last object alone (line 5). 
This path is returned by function ${\tt PUSH}$ that will be described in the next section. The last object is the head of the pushing chain, it is given by $L(N)$ and its aimed final pose is given by $p_F[L[N]]$.
$\pi_{main}$, like all paths returned by ${\tt PUSH}$, avoids collisions with the fixed  obstacles $\mathcal{S}$ as well as with objects $\mathcal{O}_s$ that have been successfully placed in the current search branch. But collisions with the remaining objects are allowed. The effects of these collisions are simulated with a physics engine and included in the sequence of states given by the path. 

Path $\pi_{main}$ can be followed simultaneously by a train of other objects lined up behind the frontal object if they have a similar or smaller footprint. 
To this end, the algorithm proceeds into placing the remaining objects in the reverse order of list $L$ (lines 8-22). Note that $L$ can also contain a single object as a special case. 
In iteration $i$, the formed chain already contains $m= N-i+1$ objects. There are various strategies to insert the $(m+1)^{th}$ object to the back of the chain. But for time efficiency, we limit the strategies to two options. The first option is to move the new object directly to the back of the chain.
This is done by first finding a {\it docking point} (line 10), which is a pose denoted by $p_{\textrm{docking,1}}$ and found by function {\tt DOCKING-POSE}, explained below. The object is then pushed alone to its docking pose through a path $\pi_1$.
The second option is to find the nearest point on path $\pi_{main}$ to the current object's pose. The nearest pose, denoted by $p_{\textrm{docking,2}}$ and returned by function {\tt NEAREST-POSE}, is obtained by using a standard $RRT^*$ search or by a heuristic that ignores other objects and projects the object's position on path $\pi_{main}$. From pose  $p_{\textrm{docking,2}}$, we compute the corresponding rendezvous pose of the frontal object (line 12). The rendezvous pose indicates where the frontal object should be moved to so that the new object joins the chain when moved to pose $p_{\textrm{docking,2}}$.
We then compare the two options, which are (a) first push the new object to the back of the partial chain then push the chain to the final pose of the frontal object, and (b) push the partial chain to that final pose and make sure the new object joins
the chain at the rendezvous point. The costs of the two options are computed and the one with the smallest cost is selected (lines 19-22). Note that this choice is not trivial, because the costs also include the transition of the end-effector, which can be initially closer to the start of option (a) or option (b). Option (b) also involves one more transition of the end-effector to bring the new object to the rendezvous point.

\begin{figure}
    \centering
        \includegraphics[width=0.23\textwidth]{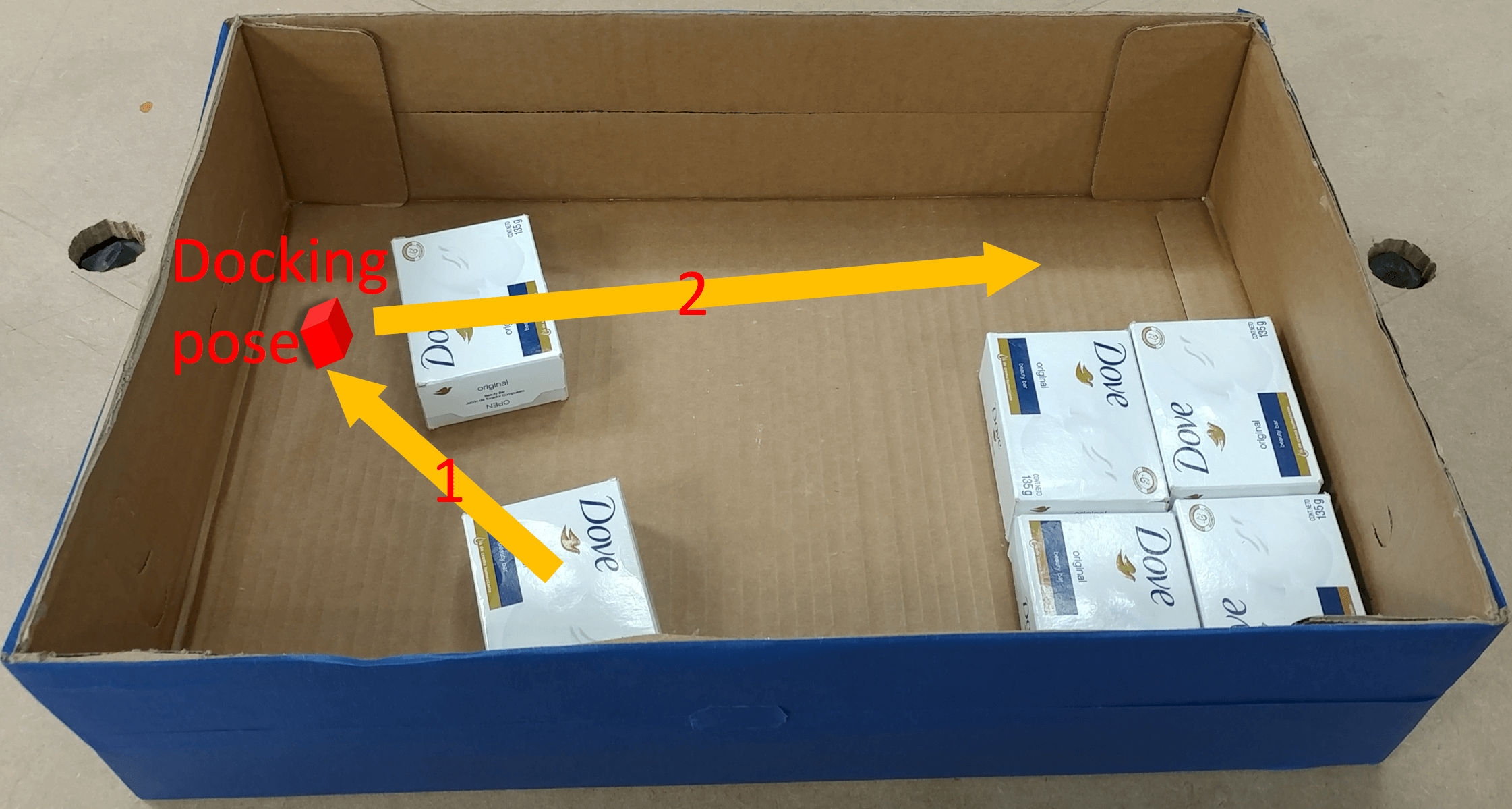}
        \includegraphics[width=0.23\textwidth]{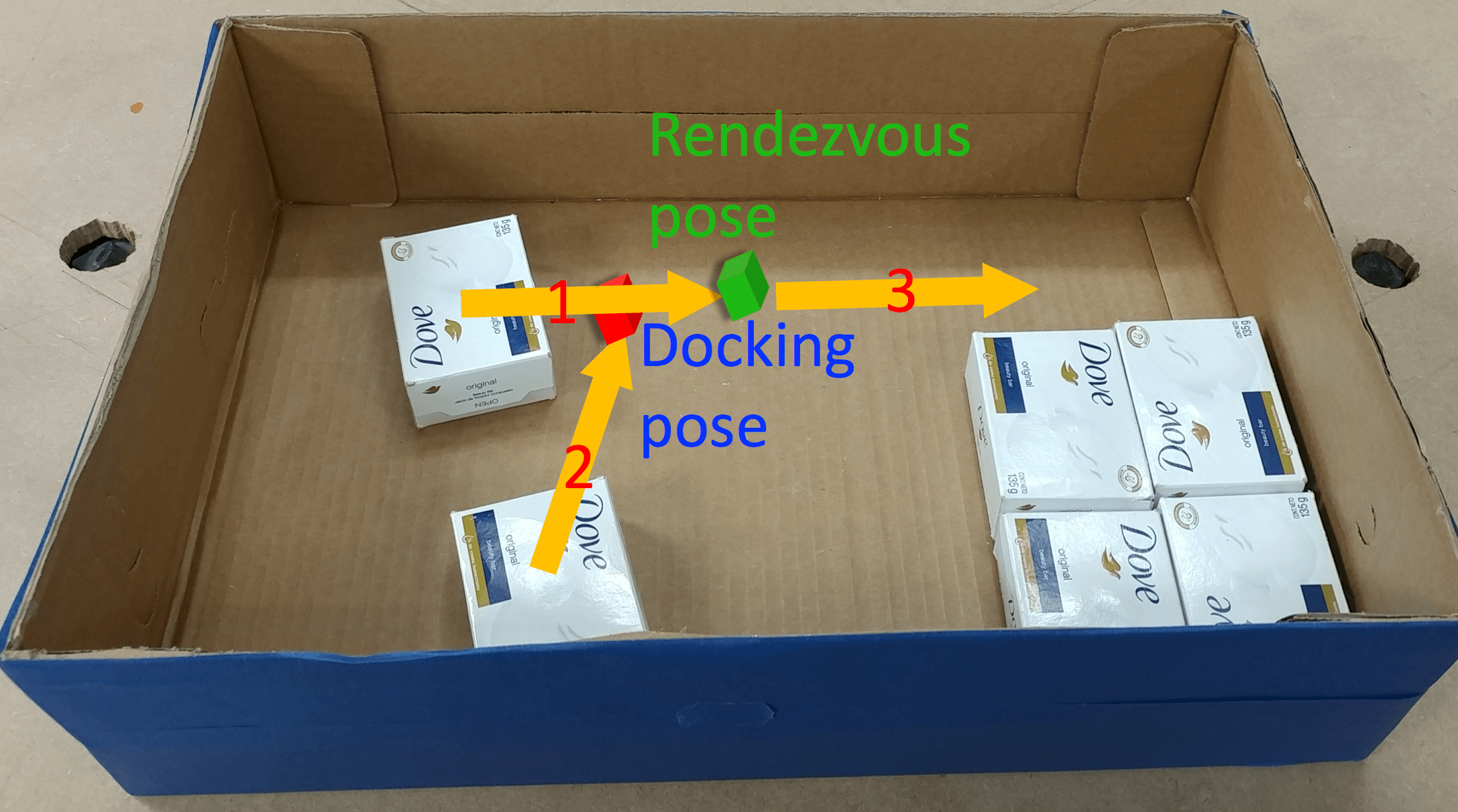}
    \caption{\small Strategies for merging paths of two objects}
    \label{docking}
\end{figure}
\begin{figure}
    \centering
        \includegraphics[width=0.23\textwidth]{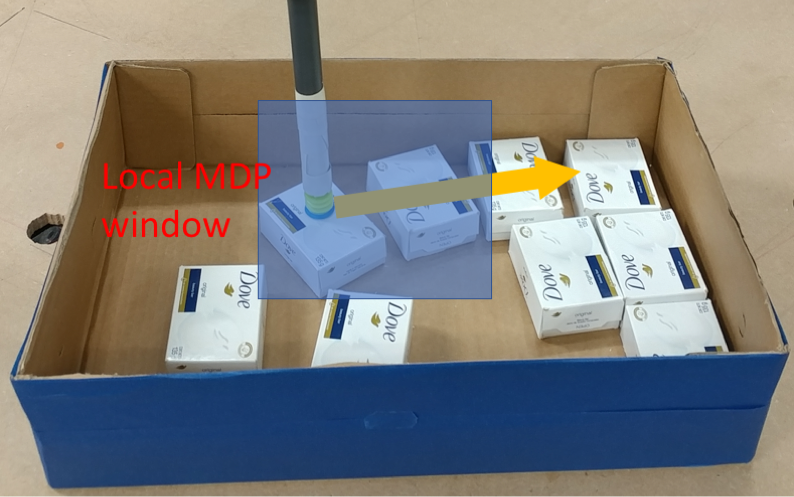}
    \caption{\small Nested pushing of objects can rearrange several objects simultaneously and reduce the trajectory of the end-effector.}
    \label{docking}
    \vspace{-0.75cm}
\end{figure}

At the end of path $\pi$, all objects of list $L$ form a chain described in the global state $(\pi(1).q,\pi(1).p)$, with different objects joining at different rendezvous points. The nested chain is pushed so that the frontal object reaches its final pose (line 23).  
The final steps (lines 25-30) consist in pushing each object individually to its final pose, while growing the set of objects $\mathcal{O}_s$ that are successfully placed (line 26). 

We now describe the procedure for computing the docking pose, given by Algorithm~\ref{alg:docking}. The docking pose is simply one that is located near the frontal object of the nested chain.  
This procedure has two modes: random and path-oriented. In the random mode, the given path is empty, a planar unitary translation vector $\vec{v}$ is randomly sampled. 
In the path-oriented mode, planar unitary translation vector $-\vec{v}$ is in the opposite direction of the provided path in pose $p_1$ of the frontal object, which ensures that the docking point is behind the frontal object. 
Pose $p_1$ is then translated in the direction of $-\vec{v}$ for a distance of $m\times 2R$ with $R$ being the radius of the largest object in the list $L$ of $m$ nested objects. 
In our experiments in the present work, we use only cuboid objects that have similar sizes, but this approach can be generalized to objects with different shapes and sizes.  
The random docking poses are useful for pushing single objects, while path-oriented docking poses are necessary for pushing trains of objects.

\begin{algorithm}[h]
\KwIn{$\pi$: pushing path, $p_I$: configuration of objects, $L$: sorted list of objects to push, $m$: number of nested objects\;}
\KwOut{Docking pose $p$ \;}
$N \leftarrow size(L)$; $i \leftarrow L[N]$ \;
\eIf{$\pi=\emptyset$}{$\vec{v}=(\frac{x}{x^2+y^2},\frac{y}{x^2+y^2},0)$, $x$ and $y$ are random\;
$p_1 = p_I[i]$\;
}
{
$x_1 = \pi(0); x_2 = \pi(\frac{1}{H})$;
$p_1 = x_1.p[i]; p_2 = x_2.p[i]$; \;
$\vec{v} \leftarrow  p_2-p_1$, $\vec{v}[3] \leftarrow 0$; $\vec{v} \leftarrow \frac{\vec{v} }{\|\vec{v} \|_2}$ \;
}
Let $R$ be the radius of the largest object in list $L$\;
return $p_1-2m\vec{v}R$\;
\caption{{\tt DOCKING-POSE}}
\label{alg:docking}
\end{algorithm}

Function {\tt RDV-POSE} of Algorithm~\ref{alg:rdv} performs the opposite function of {\tt DOCKING-POSE}, it retrieves the pose where the frontal object should be, given a docking pose, a path $\pi$, 
and a  list $L$ of $m$ nested objects to push along path $\pi$. Poses returned by {\tt RDV-POSE} and {\tt DOCKING-POSE} are translations of the frontal object's pose in $SE(2)$ that preserve its rotation, which is why $\vec{v}[3]$ is set to zero.

\begin{algorithm}[h]
\KwIn{$\pi$: pushing path, $L$: sorted list of objects to push, $m$: number of nested objects, $p_{\textrm{docking}}$: docking pose\;}
\KwOut{Rendezvous pose $p$ \;}
$N \leftarrow size(L)$; $i \leftarrow L[N]$ \;
$p_1 = p_{\textrm{docking}}$\;
Find $t\in[0,1]$ such that $\pi(t).p[i]=p_1$\;
$x_2 = \pi(t+\frac{1}{H})$; $p_2 = x_2.p[i]$\;
$\vec{v} \leftarrow  p_2-p_1$, $\vec{v}[3] \leftarrow 0$; $\vec{v} \leftarrow \frac{\vec{v} }{\|\vec{v} \|_2}$ \;
Let $R$ be the radius of the largest object in list $L$\;
return $p_1+2m\vec{v}R$\;
\caption{{\tt RDV-POSE}}
\label{alg:rdv}
\end{algorithm}
    \vspace{-0.5cm}
\subsection{Pushing Policy with Receding Horizons}
Algorithm~\ref{alg:push} returns a path $\pi$ for simultaneously pushing a set of $m$ objects nested so that the frontal object $o$ ends up in its final pose $p_F$. 
First, the algorithm attempt to sample a docking pose that is in the vicinity of the objects to push and that can be reached by the end-effector in a collision-free path $\pi_{init}$ (lines 1-5).
An $RRT^{*}$ planner is then called to find the shortest path $\pi$ for moving the frontal object to its goal (line 8). This path avoids collisions with successfully rearranged objects $\mathcal{O}_s$ and static obstacles. 
Objects are transported along $\pi$ through a sequence of highly precise pushing actions to ensure that all the pushed objects
remain on $\pi$ until they arrive at the goal. To achieve this level of accuracy, $\pi$ is divided into a sequence of $H$ small horizons. 
Each small horizon $t$ has an intermediate (waypoint) goal state $x_{target} $ that is given by state $\pi(t+1)$ (line 11). 

To reach each intermediate goal $x_{target}$ in a robust manner through a sequence of pushing actions, we formalize this problem as a local Markov Decision Process (MDP).
The state space is the cartesian product of the 3D poses of the jointly pushed objects and the end-effector. 
The 2D position part of the state space is discretized into a regular grid with a limited size, centered around the end-effector. The 1D rotation part is also discretized by dividing the unit circle into $36$ arcs. 
The action space is also discretized into a large number of small rotations and translations of the end-effector. The stochastic transition function $T$ that maps a state $s$ and action $a$ into a distribution over next
states $s'$ is learned by first fine-tuning the friction and mass coefficients of the object models inside a physics engine (Bullet), then collecting a large set of training data $\mathcal{D}=\{ (s,a,s')\}$ for random starting states $s$ and actions $a$ in the physics engine.
Since the true friction and inertial parameters of the objects are not known precisely, next states $s'$ in the data are sampled by adding small random noises to these parameters before simulating the pushing actions. 
In the real setup, the poses of the objects returned by the vision module are also subject to small errors. Therefore, the 3D poses of the objects in simulation are also perturbed by adding small random noises to the data while learning the 
transition function of the MDP model. A reward function is also defined by assigning small negative rewards to all states except for the intermediate goal states that are defined as the edges of the grid for the positions, and landmark angles for the orientations. The MDP is then solved by using the value iteration algorithm. Solving this large MDP is time consuming and took around nine hours on a single CPU. However, the MDP is solved only once and offline. The optimal policy resulting from this process is stored in a lookup table and used for all instances of rearrangement planning that involve the same objects.

\begin{algorithm}[h]
\KwIn{$(q_I,p_I)$: initial state, $o$: frontal object to push, $p_F$: final configuration of the object, $m$: number of nested objects, $\mathcal{O}_s$: set of objects that are successfully placed in their final configurations\;}
\KwOut{Path $\pi$ of the joint arm-objects configuration\;}
 \For{$i \leftarrow 0$;\ $i < max\_attempts$;\ $i \leftarrow i + 1$}
   {
      $p\leftarrow\textrm{\tt DOCKING-POSE}(\emptyset,p_I, L,m)$\;
      $\pi_{init} \leftarrow ${\tt TRANSIT}$(q_I,q(p),p_I)$\;
      \If{$\pi_{init} \neq \emptyset$}{break;}
   }
 \If{$i=max\_attempts$}
 {return $\emptyset$}
$\pi \leftarrow RRT^*(\pi_{init}(1).q,\pi_{init}(1).p,p_F, o, \mathcal{O}, \mathcal{O}_s)$\;
 \For{$t \leftarrow 0$;\ $t < 1$;\ $t \leftarrow t + \frac{1}{H}$}
   {
        $x \leftarrow \pi(t)$\;
        $x_{target} \leftarrow \pi(t+1)$\;
        \For{$i \leftarrow 0$;\ $i < horizon$;\ $i \leftarrow i + 1$}
        {
            $\delta q \leftarrow$ {\tt MDP-Policy}$(x,x_{target},m)$\;
            $x \leftarrow $ {\tt Physics-Propagation}$(x,\delta q)$\;
        }
        \If{$\|x - x_{target}\|_2 > \epsilon$ }
        {return  $\emptyset$ \;}
   }
   return  $\{\pi_{init} | \pi\}$\;
\caption{{\tt PUSH}}
\label{alg:push}
\end{algorithm}

\begin{figure}
    \centering
        \includegraphics[width=0.175\textwidth]{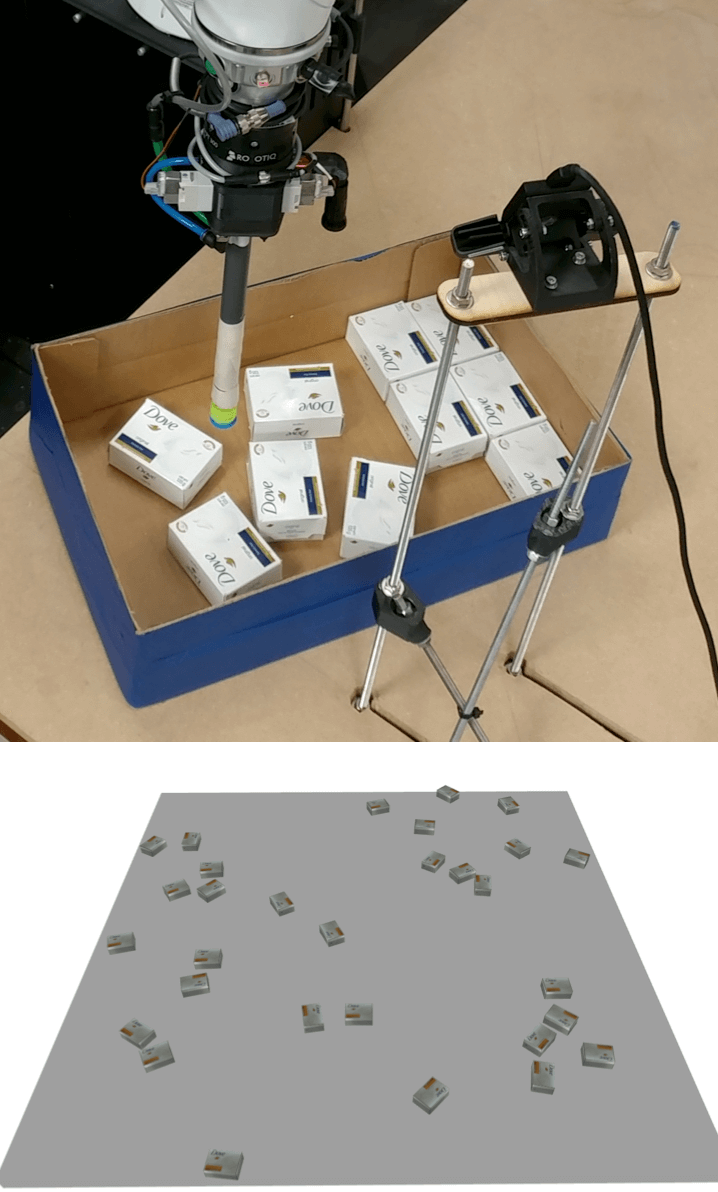}
        \includegraphics[width=0.3\textwidth]{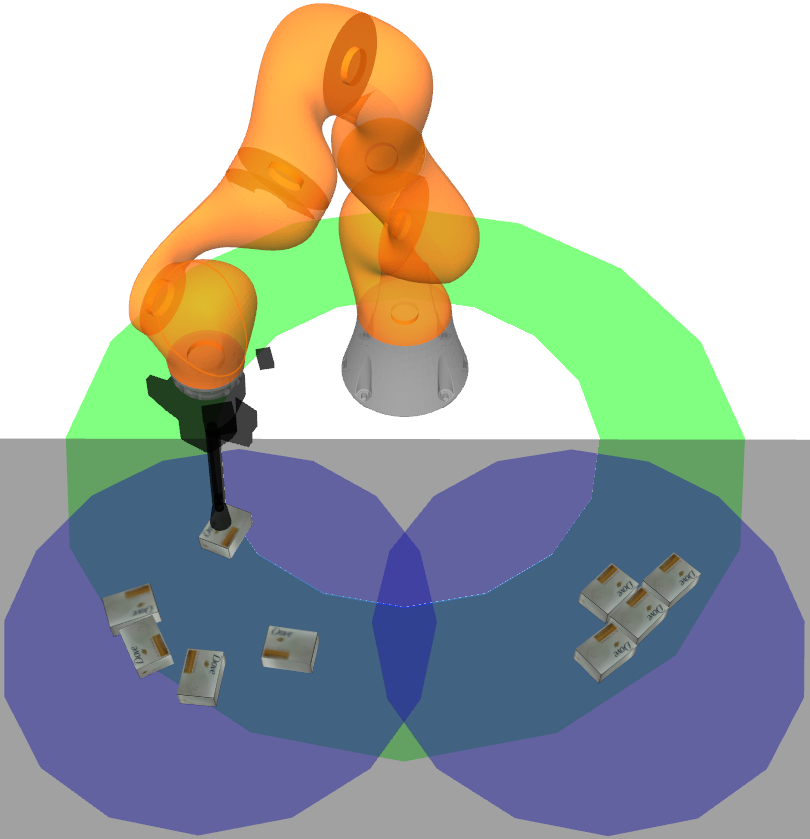}
    \caption{\small Experiments are performed on three tasks (left-down) \textit{open-space}, (right) \textit{tabletop}, and (left-up) \textit{inside-box}. The \textit{inside-box} task is performed both in simulation and with a real robot. In (right), green represents the area where the end-effector can operate freely, and blue is the field of view of the robot with two cameras.}
    \label{fig:env_setup}
\end{figure}

\section{Evaluation}
\begin{figure*}[th]
    \centering
    \begin{tabular}{ccc}
        \includegraphics[width=0.32\textwidth]{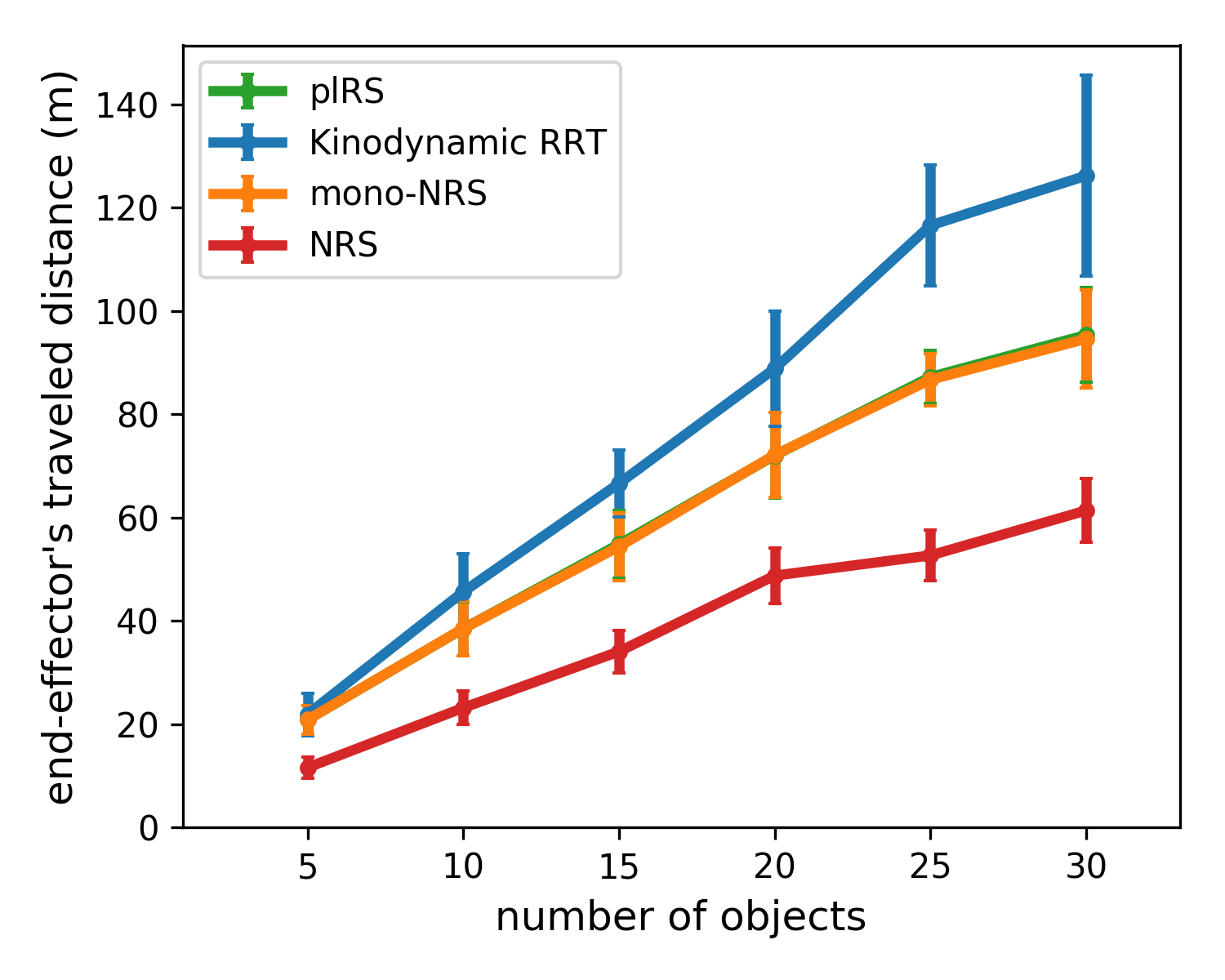} &
        \includegraphics[width=0.32\textwidth]{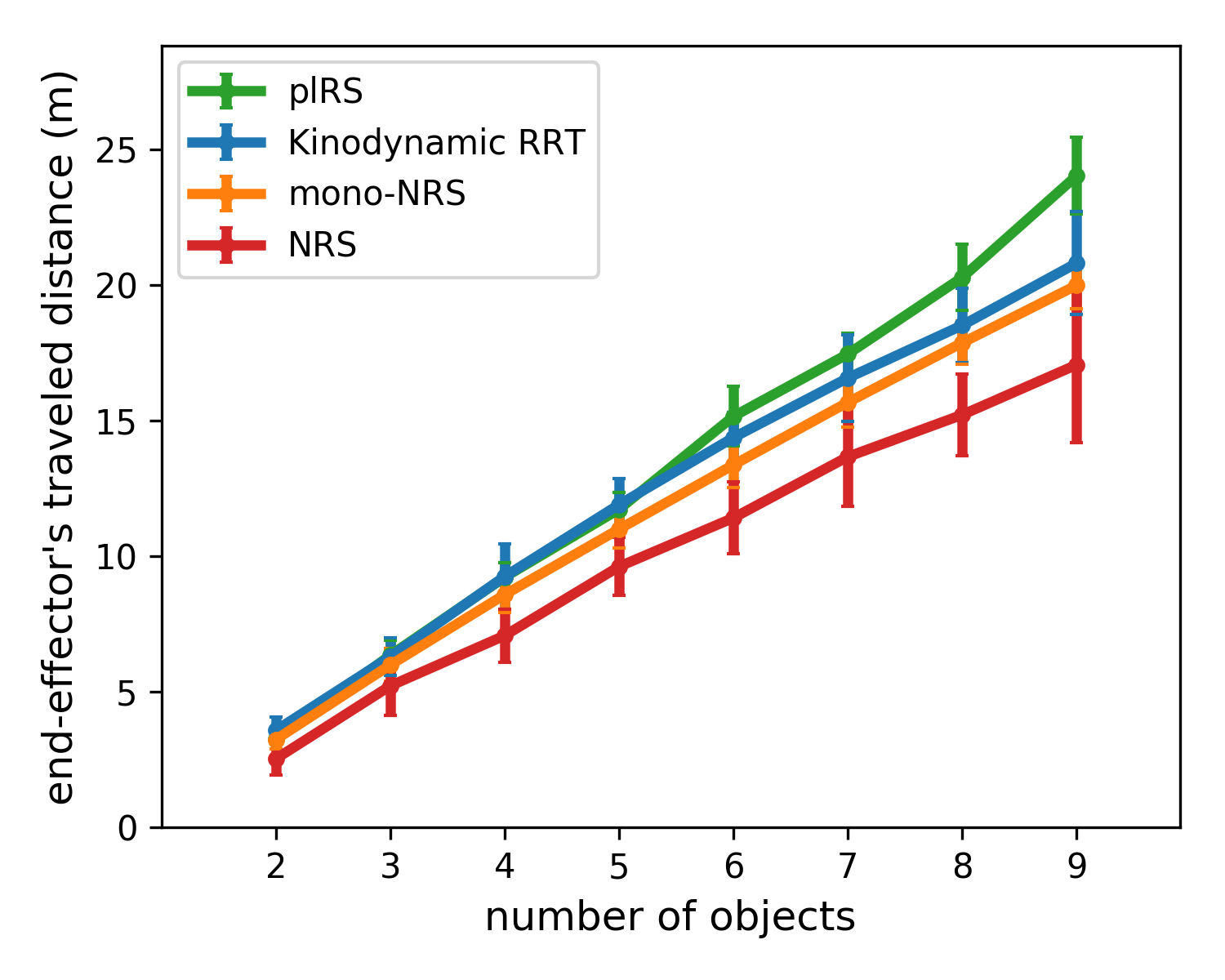} &
        \includegraphics[width=0.32\textwidth]{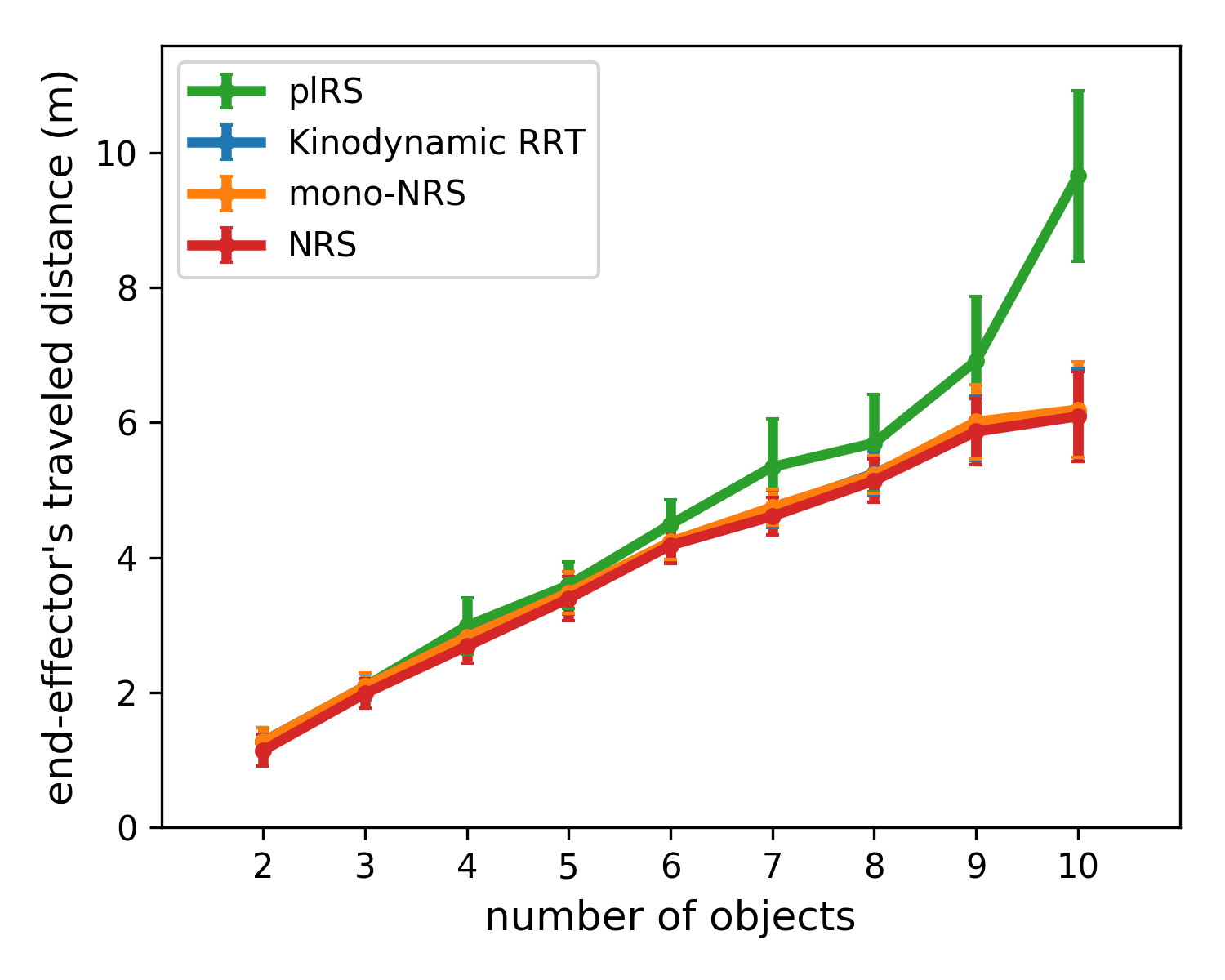} \\
	(a) \textit{open-space} &
    (b) 	\textit{tabletop} &
	(c) \textit{inside-box} 
	\end{tabular}
    \caption{Traveled distance of the end-effector as a function of the number of objects in the tasks.}
    \label{fig:eval_sim}
\end{figure*}
\begin{figure}[h]
    \centering
    \includegraphics[width=0.45\textwidth]{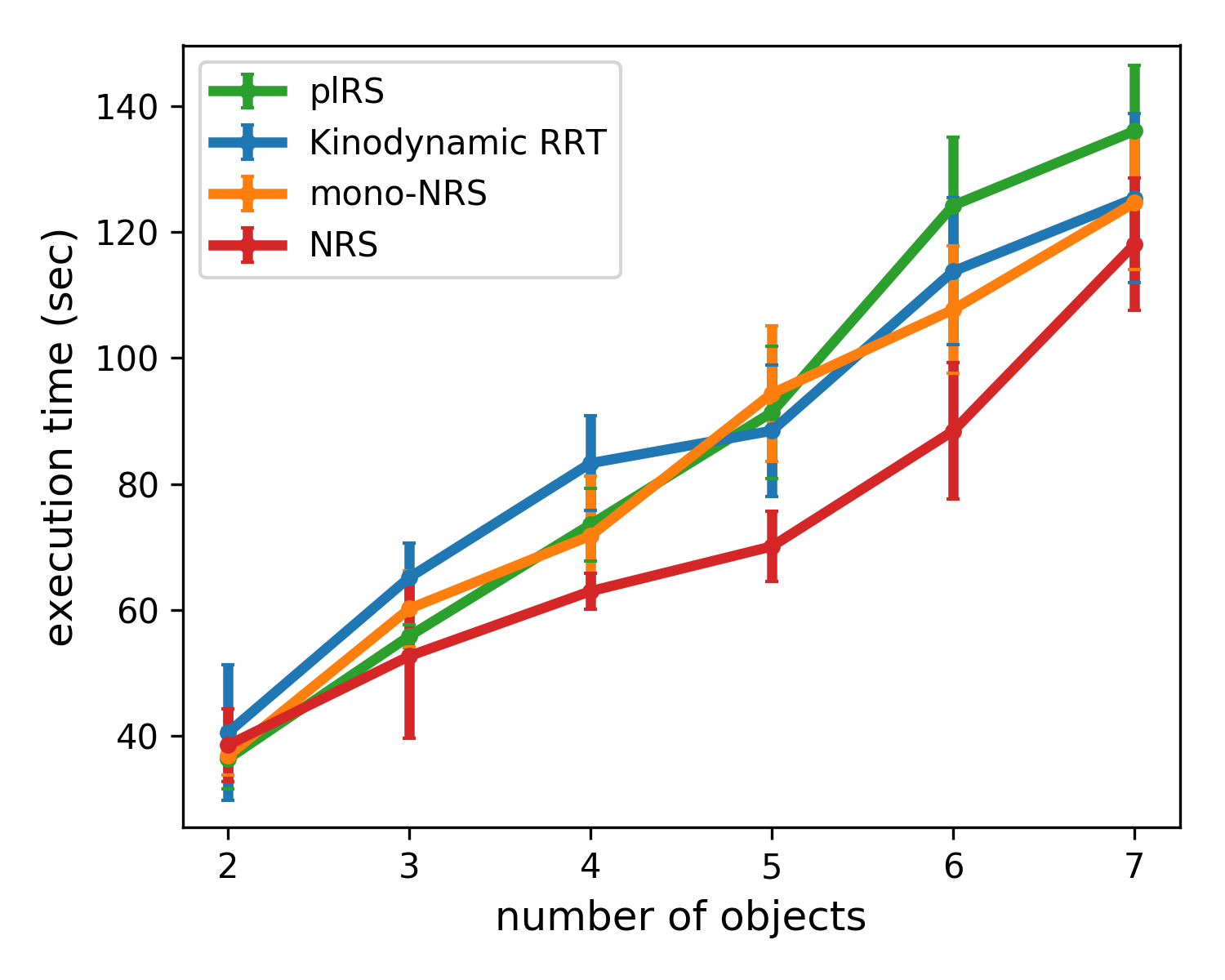}
    \caption{Execution time as a function of the number of objects in the real robot tasks.}
    \label{fig:eval_real_cost}
\end{figure}
We report here the results of our experiments for evaluating the proposed method, referred to as {\tt NRS} (Algorithm~\ref{alg:nrs}).
\subsection{Alternative approaches}
We compare the proposed method to the piecewise linear non-monotone Rearrangement Search  (plRS) method~\cite{Krontiris-RSS-15}.
The plRS method was originally proposed for pick-and-place actions, but we extend it here to handle pushing actions as well for a fair comparison. The plRS algorithm finds collision-free paths. If a path is obstructed, then the blocking objects are temporarily moved off the way to the nearest pose in the free space. At a high level, plRS is similar to our method in the sense that it also uses tree search with backtracking to find the order in which the object are moved. To assess the efficiency of the MDP policy for pushing objects, we substitute function {\tt PUSH} in our algorithm with the Kinodynamic RRT procedure presented in~\cite{king2015} and compare the two methods as well. Kinodynamic RRT was used in~\cite{king2015} for nonprehensile rearrangement. 

We also compare to a variant of our method, referred to as {\tt mono-NRS}, which is identical to {\tt NRS} except that the number of pushed objects in a given path is limited to one. This is equivalent to replacing the powerset in line 4 of Algorithm~\ref{alg:nrs} with a set of mono-object sets.

\subsection{Experiment Setup}
We used the methods discussed above to solve four rearrangement types of tasks, three in simulation and one with a real robot.
The simulation tasks are \textit{open-space}, \textit{tabletop}, and \textit{inside-box}.
The \textit{open-space} setup is a 2m$\times$2m workspace without obstacles.
The \textit{tabletop} setup is a more realistic environment constrained by the robot's configuration space and its field of view as shown in Figure~\ref{fig:env_setup}.
The \textit{inside-box} setup is another realistic environment with a tight free space.
For each task, $20$ different scenes are generated by sampling random poses of various numbers of objects. The reported results are averages of these independent runs. 
In the \textit{open-space} task, the targeted final poses are arbitrary. In \textit{tabletop}, objects are pushed from the left side and packed on the right side. In \textit{inside-box}, the targeted final poses are assigned to objects based on their initial poses, each object is assigned to the final pose that is nearest to its initial pose.

The real robot experiments are performed on the setup shown in Figure~\ref{fig:problem_bins}, using 9.5cm$\times$6.5cm$\times$3.5cm cuboid objects inside a 49cm$\times$33cm cardboard box.
We randomly generated the initial collision-free poses of the objects in each round, and used the same initial poses for all methods. Poses of objects are then estimated by segmenting the point cloud of the scene and performing PCA on the clusters. The final arrangement poses are those shown in Figure \ref{fig:problem_bins} (right). The experiments are performed on bins with two to seven objects. Each experiment for each method is repeated four times. The total number of tested real scenes is $96$ $(6\times4\times4)$.

\subsection{Results}

\begin{table}[h]
\centering
\begin{tabular}{|l|c|c|c|c|}
\cline{2-3}
\multicolumn{1}{c|}{}
&  Position & Rotation \\ \hline
plRS \cite{Krontiris-RSS-15} &  $85.2$\%$ (\pm10.9)$ & $82.4$\%$ (\pm12.2)$ \\
Kinodynamic RRT \cite{king2015} & $85.2$\%$ (\pm10.9)$ & $76.9$\%$ (\pm14.3)$\\
mono-NRS & $85.2$\%$ (\pm03)$ & $81.5$\%$ (\pm06)$\\
NRS & $88.0$\%$ (\pm05.6)$ & $86.1$\%$ (\pm08.2)$\\ 
\hline
\end{tabular}

\caption{Success rates in the real robot experiment.}
\label{tbl:eval_real_succ}
\end{table}

The time limit for all methods is set to $10$ minutes per task. 
Within this limit, all methods achieved a $100\%$ success rate in simulation. A successful object placement is defined as one where the final reached pose is within $1cm$ and $10^{\circ}$ of the final desired pose.

Figure~\ref{fig:eval_sim} shows the total traveled distance of the end-effector as a function of the number objects used in the simulated rearrangement task. 
The results demonstrate that the nested-push approach effectively reduces the end-effector's movement by moving multiple objects together. It also seems like the advantage of this method becomes more pronounced as more objects are considered. In fact, finding objects to push together becomes easier in the presence of multiple objects. On the other hand, when too many objects are cramped inside a small space, it becomes more difficult to successfully push more than one object at a time, which is the reason why the advantage of  {\tt NRS} is clearer in task \textit{open-space}. Because collisions are allowed in the {\tt mono-NRS} method, objects are sometimes accidentally pushed to their targeted final poses, which contributes to reducing the cost of that method in \textit{inside-box}. Accidental placements are less likely to occur in large open spaces, such as \textit{open-space}


The threshold for success in the real robotic experiments, reported in Table~\ref{tbl:eval_real_succ}, is defined as $2cm$ for translation errors and $15^\circ$ for rotation errors. All methods achieved a reasonable success rate, with no significant difference between the methods. Execution time, however, is significantly smaller when {\tt NRS} is used, as shown in Figure~\ref{fig:eval_real_cost}, which confirms the simulation results.



\bibliographystyle{IEEEtran}
\bibliography{references}

\end{document}